\documentclass[runningheads]{llncs}
\usepackage{graphicx}
\usepackage{graphicx}
\usepackage{booktabs}
\usepackage{multicol}
\usepackage{multirow}
\usepackage{amsmath}
\usepackage{amssymb}
\usepackage{bbding}
\usepackage{subfigure}
\usepackage{colortbl}
\usepackage{pifont}
\usepackage{CJKutf8}

\newcommand{\ourmethod}{\textsc{HanoiT}}

\begin{document}
\begin{CJK*}{UTF8}{gbsn}
\title{\ourmethod{}: Enhancing Context-aware Translation via Selective Context}


\author{
  Jian Yang\textsuperscript{\rm 1}, 
  Yuwei Yin\textsuperscript{\rm 2}, 
  Shuming Ma\textsuperscript{\rm 3},
  Liqun Yang\textsuperscript{\rm 1 \thanks{\ Corresponding author.}},
  Hongcheng Guo\textsuperscript{\rm 1},
  Haoyang Huang\textsuperscript{\rm 3}, 
  Dongdong Zhang\textsuperscript{\rm 3},
  Yutao Zeng\textsuperscript{\rm 1}, 
  Zhoujun Li\textsuperscript{\rm 1}, 
  Furu Wei\textsuperscript{\rm 2} \\
  \textsuperscript{\rm 1}State Key Lab of Software Development Environment, Beihang University \\ 
  \textsuperscript{\rm 2}The University of Hong Kong; 
  \textsuperscript{\rm 3}Microsoft Research Asia \\
  \{jiaya, lqyang, hongchengguo, zengyutao, lizj\}@buaa.edu.cn;\\ yuweiyin@hku.hk; \\\{shumma, haohua, dozhang, fuwei\}@microsoft.com 
}

\institute{}

\maketitle


\begin{abstract}
Context-aware neural machine translation aims to use the document-level context to improve translation quality. However, not all words in the context are helpful. The irrelevant or trivial words may bring some noise and distract the model from learning the relationship between the current sentence and auxiliary context.
To mitigate this problem, we propose a novel end-to-end encoder-decoder model with a layer-wise selection mechanism to sift and refine the long document context.
To verify the effectiveness of our method, extensive experiments and extra quantitative analysis are conducted on four document-level machine translation benchmarks. The experimental results demonstrate that our model significantly outperforms previous models on all datasets via the soft selection mechanism.

\end{abstract}

\section{Introduction}

Recently, neural machine translation (NMT) based on the encoder-decoder framework has achieved state-of-the-art performance on the sentence-level translation \cite{Seq2Seq,Seq2Seq2,GNMT,deliberation_network,Transformer,ConvS2S,multipass_decoder,LightConv,insertion_transformer}. However, the sentence-level translation solely considers single isolated sentence in the document and ignores the semantic knowledge and relationship among them, causing difficulty in dealing with the discourse phenomenon such as lexis, ellipsis, and lexical cohesion \cite{HAN,CADec}.

To model the document-level context, there are two main context-aware neural machine translation schemes. One approach introduces an additional context encoder to construct dual-encoder structure, which encodes the current source sentence and context sentences separately and then incorporates them via the gate mechanism \cite{larger_context,Cross_sentence_context,Discourse_phenomena,Anaphora_Resolution,selective_attention,source_side_context_aware}.
The other one directly concatenates the current source sentence and context sentences as a whole input to the standard Transformer architecture, though the input sequence might be quite long \cite{Extended_Context,Evaluating_Discourse_Phenomena,Extended_Context_2,Extended_context_3}.
The previous works \cite{Extended_Context_2,Flat_Transformer} conclude  that the Transformer model has the capability to capture long-range dependencies, where the self-attention mechanism enables the simple concatenation method to have competitive performance with multi-encoder approaches.

\begin{figure}[t]
\begin{center}
    \includegraphics[width=0.8\textwidth]{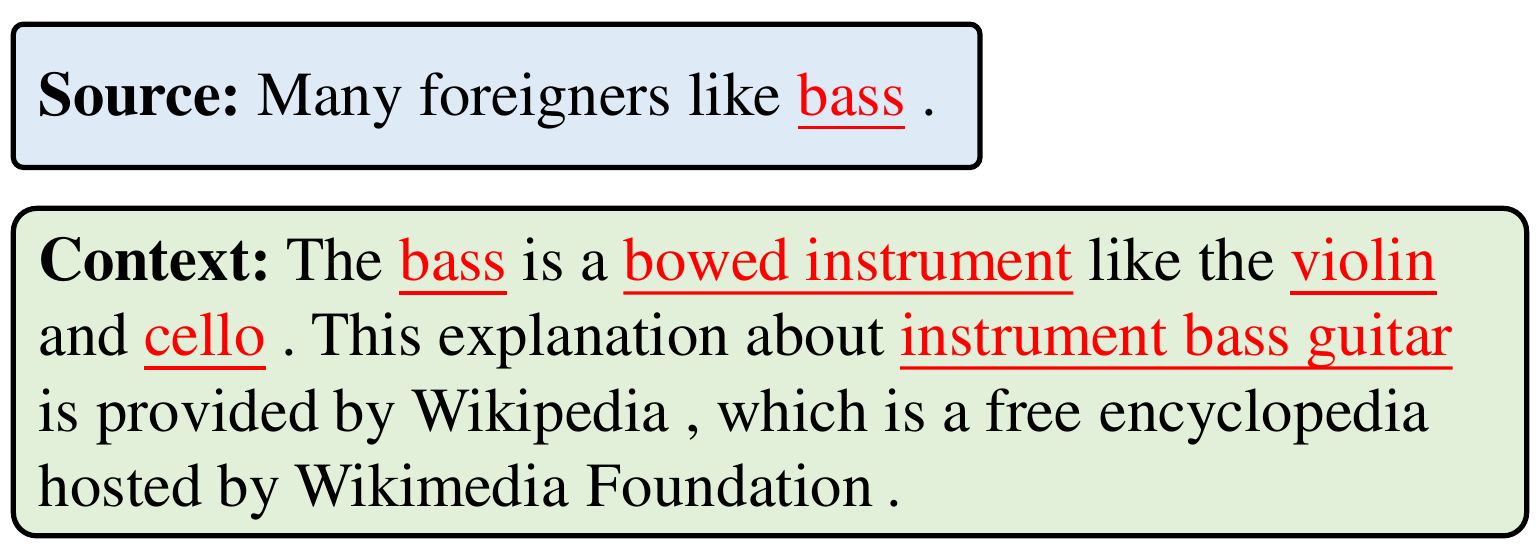}
    \caption{An example of the source and the context sentence. Above is a source sentence to be translated, and below is its context in the same document. The underlined words are useful to disambiguate the source sentence, while the rest is less important.}
    \vspace{-15pt}
    \label{intro}
\end{center}
\end{figure}

Most aforementioned previous methods use the whole context sentences and assume that all words in the context have a positive effect on the final translation. Despite the benefits of part of the context, not all context words are useful to the current translation. In Figure \ref{intro}, the underlined words provide supplementary information for disambiguation, while the others are less important. The irrelevant words may bring some noise and redundant content, increasing the difficulty for the model to learn the relationship between the context and the translation. Therefore, these useless words should be discarded so that the model can focus on the relevant information of the current sentence.

In this work, we propose an end-to-end model to translate the source document based on layer-wise context selection over encoder.
In our model, the context is concatenated with current source sentence as external knowledge to be fed into the unified self-attention, where they are precisely selected among multiple layers to gradually discard useless information. The criteria on context selection is based on context-to-source attention score which are recursively calculated layer-by-layer. Ultimately, the context on the top layer is expected to be the most useful knowledge to help current source sentence translation. The architecture of our model looks like a Tower of \textbf{Hanoi} over the \textbf{T}ransformer structure (\ourmethod{}). Our proposed model captures all context words at the bottom layer and focuses more on the essential parts at the top layer via the soft selection mechanism.

To verify the effectiveness of our method, we conduct main experiments and quantitative analysis on four popular benchmarks, including IWSLT-2017, NC-2016, WMT-2019, and Europarl datasets.
Experimental results demonstrate that our method significantly outperforms previous baselines on these four popular benchmarks and can be further enhanced by the sequence-to-sequence pretrained model, such as BART \cite{bart}.
Analytic experiments and attention visualization illustrate our proposed selection mechanism for avoiding the negative interference introduced by noisy context words and focusing more on advantageous context pieces.

\section{Our Approach}
\label{Our Approach}
In this section, we will describe the architecture of our \ourmethod{}, and apply \ourmethod{} to context-aware machine translation.

\subsection{Problem Statement}
Formally, let $X = \{x^{(1)},..,x^{(k)},..,x^{(K)}\}$ denote a source language document composed of $K$ source sentences, and $Y = \{y^{(1)},..,y^{(k)},..,y^{(K)}\}$ is the corresponding target language document. $\{x^{(k)},y^{(k)}\}$ forms a parallel sentence, where $x^{(k)}$ denotes the current source sentence and $y^{(k)}$ is the translation of $x^{(k)}$. $X_{<k}=\{x^{(1)},..,x^{(k-1)}\}$ denotes the historical context and $X_{>k}=\{x^{(k+1)},..,x^{(K)}\}$ represents the future context. Given the current source sentence $x^{(k)}$, the historical context $X_{<k}$, and the 
future context $X_{>k}$, the translation probability is given by:
\begin{MiddleEquation}
\begin{align}
    \begin{split}
       P(y^{(k)}|X;\theta)=\prod_{i=1}^{N}P(y^{(k)}_{i}|X;y^{(k)}_{<i};\theta)
    \end{split}
    \label{source_lm}
\end{align}
\end{MiddleEquation}where $y^{(k)}_i$ is the $i^{th}$ word of the $k^{th}$ target sentence and $y^{(k)}_{<i}$ are the previously generated words of the target sentence $y^{(k)}$ before $i^{th}$ position. $y^{(k)}$ has $N$ words.
In this work, we use one previous and one next sentence as the context.

\begin{figure*}[t]
\begin{center}
    \includegraphics[width=0.95\textwidth]{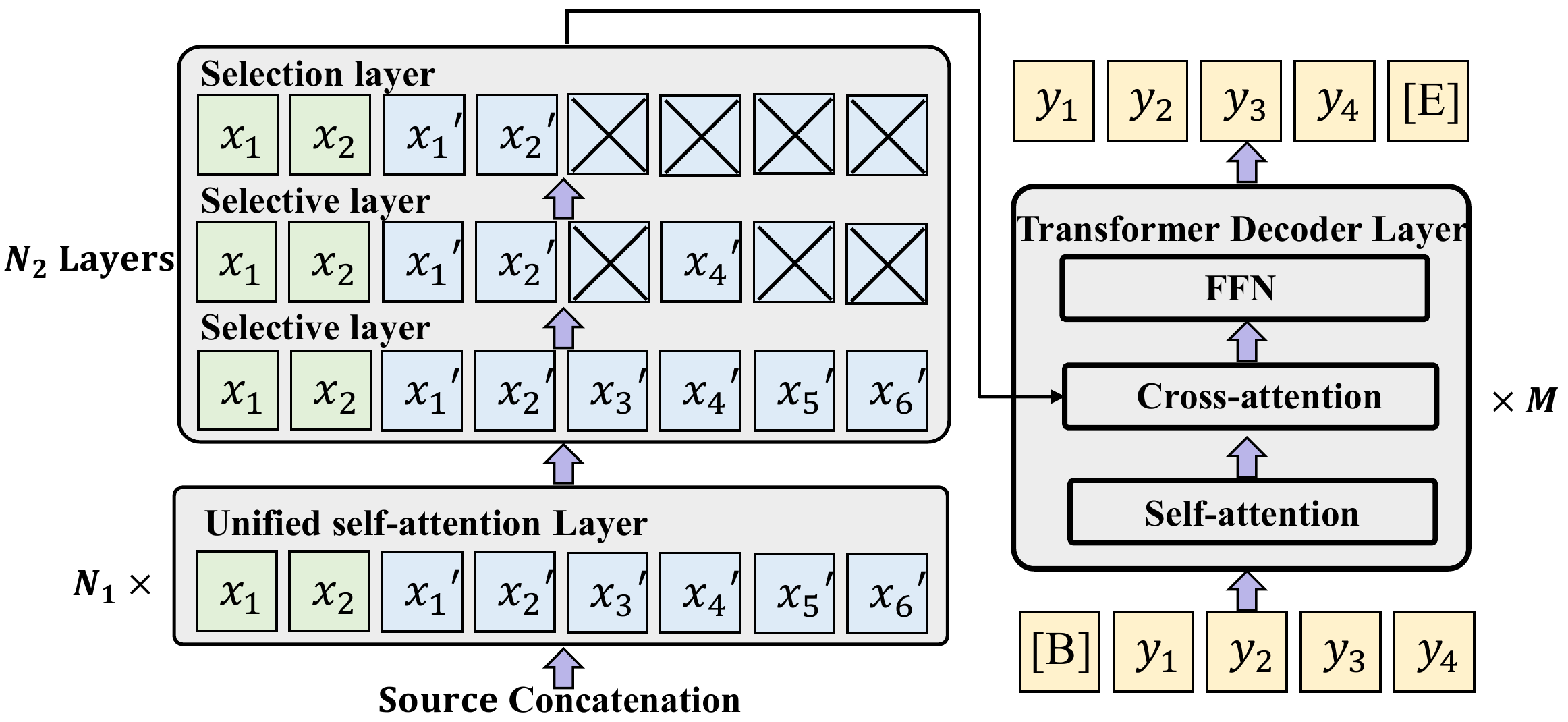}
    \caption{Overview of our proposed model \ourmethod{}. For simplicity, layer normalization and other components of the Transformer architecture are omitted in the picture. Cross symbols denote dropped words.
    $(x_1,x_2)$ is the current source sentence and $(x_1',x_2',x_3',x_4',x_5',x_6')$ is the source context. $N_1$ and $N_2$ denote the number of unified self-attention layers and selection layers. $(x_1',x_2',x_3',x_4',x_5',x_6') \rightarrow (x_1',x_2',x_4') \rightarrow (x_1',x_2')$ is the selective procedure, where important words are selected gradually by multiple selection layers.}
    \label{model}
\end{center}
\end{figure*}

\subsection{\ourmethod{}}
 \label{sourec-context}
Figure~\ref{model} shows the overall structure of our \ourmethod{} model. At the bottom of the encoder, it models the concatenation of the source sentence and the context with unified self-attention layers. At the top of the encoder, it gradually selects the context words according to the attention weights.

\paragraph{Embedding}
We use the segment embedding to distinguish the current sentence, source, and target context sentences. In Figure \ref{model}, we concatenate the current sentence and the source context as a whole. To model the positions of the different parts, we also reset the positions of the current source sentence and source context sentences. Therefore, the final embedding of the input words is the sum of the word embedding, position embedding, and segment embedding, which can be described as:
\begin{align}
    \begin{split}
       E = E_{w} + E_{p} + E_{s} 
    \end{split}
\end{align}where $E_w$ is the word embedding, $E_s$ is the segment embedding from the learned parameter matrix, and $E_p$ is the position embedding. 

\paragraph{Encoder}
Since the inputs of context-aware neural machine translation are composed of several sentences, we build our model based on the multi-head attention to capture long-range dependencies and compute the representation of the document-level context. Our encoder consists of two groups of layers: unified self-attention layers and selection layers. The unified self-attention layers is to compute a joint representation of the source sentence and the context, while the selection layer is to select the context for the next layer.

\paragraph{Unified Self-attention Layer}
Given the concatenation of the source sentence and the source context, we obtain the document representation $s^0=\{s_1^{0},..,s_{p}^{0},..,,s_m^{0}\}$ after the embedding layer, where $p$ is the length of $x^{(k)}$ and $m$ is the length of source concatenation. Then, we feed the $s^0$ into $N_1$ unified self-attention layers to compute their representations.
\begin{MiddleEquation}
\begin{align}
    \begin{split}
       s^l = \text{FFN}(\text{MultiHeadAttn}(s^{l-1};\theta_{N_1}))
    \end{split}
\end{align}
\end{MiddleEquation}where the $l$ is the number of the unified self-attention layer and $l \in [1,N_1]$. 
    
\paragraph{selection layer}
After $N_1$ unified self-attention layers, we get representations of source concatenation $s^{N_1}=\{s_1^{N_1},..,s_p^{N_1},..,s_m^{N_1}\}$, which can be used to select important context words.
In the selection layer, we apply multi-head attention to $s^{N_1}$, and then average attention scores across different heads, which can be described as below:
\begin{MiddleEquation}
\begin{align}
    \begin{split}
       a_{i,j}= \frac{1}{h}\sum_{1 \le i \le h}\text{MultiHeadAttn}(s^{N_1})
    \end{split}
\end{align}
\end{MiddleEquation}where $h$ is the number of attention heads. $a_{i,j}$ represents the average attention score between the $i^{th}$ token and the $j^{th}$ token. 

Then, we calculate the average attention score between $i^{th}$ word and other tokens in the source current sentence $x^{(k)}$:
\begin{MiddleEquation}
\begin{align}
    \begin{split}
       a_{i,\not=i}= \frac{1}{p}\sum_{j \in [1,p], j \not= i}a_{i,j}
    \end{split}
\end{align}
\end{MiddleEquation}where $a_{i,\not=i}$ represents the average correlation between $i^{th}$ word and the other words, and $p$ is the number of tokens in the source sentence.


In order to decide which context words should be selected, we compute the correlation scores $s$ between each context word and the whole source sentence. For the $k^{th}$ context word, we count how many words in the current sentence have a higher attention score with it compared to the average attention score $a_{i,\not=i}$:
\begin{MiddleEquation}
\begin{align}
    \begin{split}
       s_{k}= \sum_{i\in [p+1,m]}\delta_{a_{i,k} \ge a_{i,\not= i}}
    \end{split}
    \label{percent}
\end{align} 
\end{MiddleEquation}where $\delta_{a_{i,k} \ge a_{i, \neq i}}$ equals $1$ if $a_{i,k} \ge a_{i,\not= i}$ else $0$, $p$ is the number of tokens in the source sentence, and $m$ is the total number of tokens in the concatenation of the source sentence and the source context.

Finally, we can select the context words with top correlation scores $s_{k}$. We use $v_k$ to denote whether the $k^{th}$ word is selected:
\begin{MiddleEquation}
\begin{align}
    \begin{split}
       v_k = \delta_{s_k \ge q * p}
    \end{split} 
\end{align}
\end{MiddleEquation}where $\delta_{s_k \ge q * p}$ equals to $1$ if $s_k \ge q * p$ else $0$, $p$ is the number of tokens in the source sentence, and $q \le 1$ is a hyper-parameter to control the percentage of the selective context. In this work, we set $q \in [0.1, 0.5]$ according to the performance in the validation set.

\begin{figure*}[t]
\begin{center}
    \includegraphics[width=0.95\textwidth]{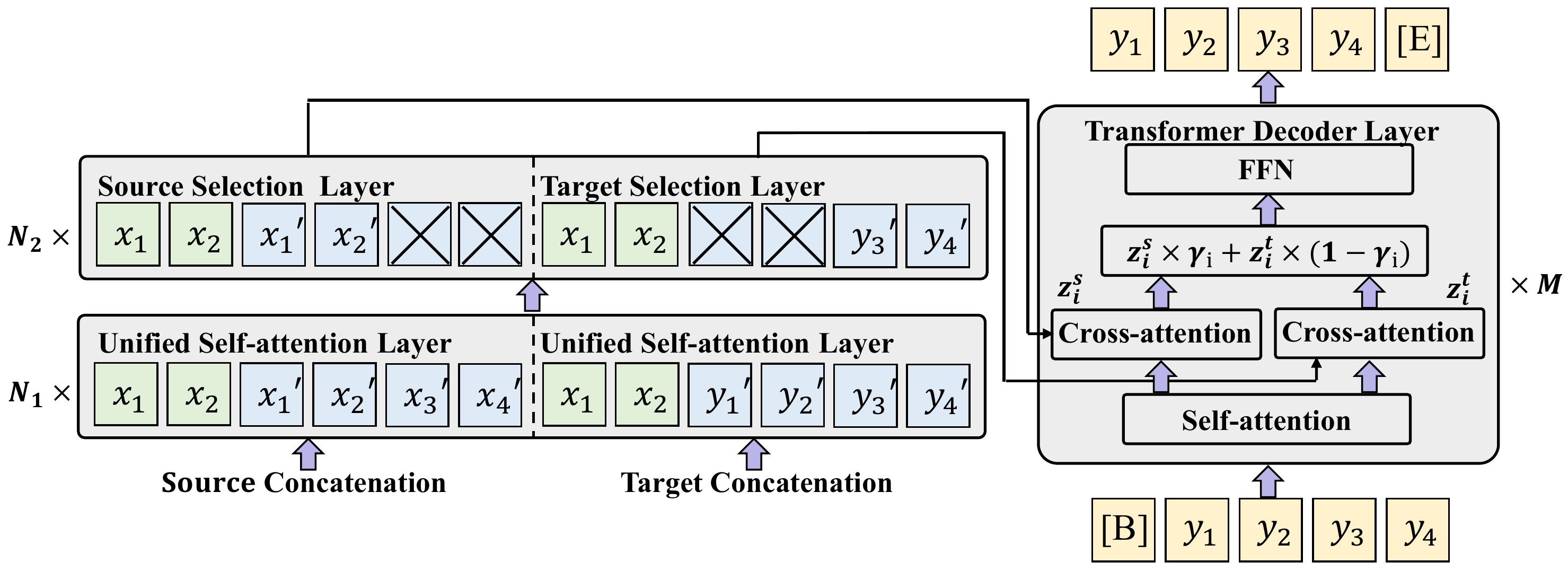}
    \caption{Overview of the extended \ourmethod{} to integrate the bilingual context. Cross symbols denote masked words. Source concatenation consists of the current source sentence $(x_1,x_2)$ and source context $(x_1',x_2',x_3',x_4')$. Target concatenation is composed of the source sentence $(x_1,x_2)$ and the target context $(y_1',y_2',y_3',y_4')$. Then the source and target selective concatenations are incorporated by the gate mechanism to predict the final translation. }
    \label{model1}
\end{center}
\end{figure*}

\paragraph{Decoder} The source selective concatenation $s^{N_2}=\{s_1^{N_2},..,s_p^{N_2},..,s_{m_1}^{N_2}\}$ is fed into the standard Transformer decoder to predict the final translation. 

\subsection{Bi-lingual Context Integration}

Section \ref{sourec-context} only considers the mono-lingual context, i.e. the source context. In practice, when translating a document, we can also obtain the target context by sentence-level translating the document before context-aware translation~\cite{CADec}. In this section, we extend our \ourmethod{} model to integrate the bi-lingual context, i.e. the source context and the target context.

Formally, let $X = \{x^{(1)},..,x^{(k)},..,x^{(K)}\}$ denote a source language document composed of $K$ source sentences and $Y = \{y^{(1)},..,y^{(k)},..,y^{(K)}\}$ denotes the sentence-level translation of $X$.
$X_{<k}$ is the historical source context and $X_{>k}$ is the future source context. Similarity, we denote historical target context $\{y^{(1)},..,y^{(k-1)}\}$ as $Y_{<k}$ and future target context $\{y^{(k+1)},..,y^{(K)}\}$ as $Y_{>k}$. 
We model the translation probability that is conditioned on the bi-lingual source context $X_{\not=k}$ and target context $Y_{\not=k}$ as:
\begin{MiddleEquation}
\begin{align}
    \begin{split}
       P(y^{(k)}|X;\theta)=\prod_{i=1}^{N}P(y^{(k)}_{i}|X;Y_{\not=k};y^{(k)}_{<i};\theta)
    \end{split}
    \label{target_lm}
\end{align} 
\end{MiddleEquation}where $y^{(k)}_i$ is the $i^{th}$ word of the $k^{th}$ target sentence and $y^{(k)}_{<i}$ are the previously generated words of the target sentence $y^{(k)}$ before $i^{th}$ position. 

\paragraph{Encoder}
As shown in Figure \ref{model1}, the current source sentence and the source context are merged as the source concatenation. Besides, the current source sentence and the target context are also merged as the target concatenation. Both concatenations are fed into unified self-attention and selection layers to compute representations of source concatenation $s^{N_2}$ and target concatenation $t^{N_2}$.

\paragraph{Decoder}
With the above encoder, we obtain the representations of the selective source concatenation $s^{N_2}= \{s_1^{N_2},..,s_p^{N_2},..,s_{m_1}^{N_2} \}$ and the selective target concatenation $t^{N_2}=\{t_1^{N_2},..,t_p^{N_2},..,t_{n_1}^{N_2}\}$, where $m_1$ and $n_1$ are lengths of selective source and target concatenation. Given both selective concatenations, we deploy the multi-head attention by two attention components. Using query, key, value parameters $(W^Q_{s},W^V_{s},W^K_{s})$, the decoder gets the hidden state $z_{i}^s$. Similarly, another hidden state $z_{i}^t$ is generated by the additional attention component with parameters $(W^Q_{t},W^V_{t},W^K_{t})$.
Considering the previous insight \cite{Marcin} that the gate network is a good component for bi-lingual context setting, we employ the gate mechanism to incorporate the source and target context.
\paragraph{Gate Mechanism}
Given the $i^{th}$ hidden states $z^{s}_{i}$ and $z^{t}_{i}$, the gate mechanism can be described as:
\begin{MiddleEquation}
\begin{align}
    \begin{split}
       \gamma_i = c\sigma(W_sz_{i}^{s}+U_tz_{i}^{t}+b)
    \end{split}
\end{align}
\end{MiddleEquation}where $W_s$ and $U_t$ are parameters matrices and $b$ is a bias. 
$c \in [0,1]$ is a hyper-parameter to control range of the gate weight. $\sigma(\cdot)$ is the sigmoid function.

\begin{MiddleEquation}
\begin{align}
    \begin{split}
       z_{i} = (1-\gamma_i) z_{i}^{s} + \gamma_i z_{i}^{t}
    \end{split}
\end{align}
\end{MiddleEquation}where $z_{i}$ is the $i^{th}$ decoder final hidden state derived from the source context and target context. 

\subsection{Training}
Given the mono-lingual context only, the training objective is a cross-entropy loss function on the top of Equation \ref{source_lm}. The objective $\mathcal{L}_{m}$ is written as:
\begin{MiddleEquation}
\begin{align}
\begin{split}
    \mathcal{L}_{m} = -\sum_{X,y^{(k)} \in D}\log{P_{\theta}(y^{(k)}|X)}
    \label{eq_mono}
\end{split}
\end{align}
\end{MiddleEquation}where $\theta$ are model parameters. 

Considering the bi-lingual context, the training objective $\mathcal{L}_{b}$ is calculated as:
\begin{MiddleEquation}
\begin{align}
\begin{split}
    \mathcal{L}_{b} = -\sum_{X,y^{(k)},Y_{\not=k} \in D}\log{P_{\theta}(y^{(k)}|X,Y_{\not=K})}
\end{split}
\end{align}
\end{MiddleEquation}where $\theta$ are model parameters. 

The quality of the target context depends on the sentence-level translation model, which may bring additional errors. To reduce the possible harm by these errors and make the training stable, our model optimizes a combination of the mono-lingual objective $\mathcal{L}_{m}$ and the bi-lingual objective $\mathcal{L}_{b}$:
\begin{MiddleEquation}
\begin{align}
\begin{split}
    \mathcal{L}_{all} = \alpha  \mathcal{L}_{m} + (1-\alpha) \mathcal{L}_{b}
    \label{eq_bi}
\end{split}
\end{align}
\end{MiddleEquation}where $\alpha$ is a scaling factor to balance two objectives between $\mathcal{L}_{m}$ and $\mathcal{L}_{b}$. We find when the value of $\alpha$ equals 0.5, our model gets the optimal performance by balancing two objectives. We adopt Equation~\ref{eq_mono} to train the model with mono-lingual context, and Equation~\ref{eq_bi} to train the model with bi-lingual context.

\section{Experiments}
\label{Experiments}

To prove the efficiency of our method, we conduct experiments on four public benchmarks. 

\begin{table*}[t]
\begin{center}
\caption{\label{one-pass}  Sentence-level evaluation results on four tasks with BLEU\% metric using the source context. Bold numbers denote the best BLEU points. RNN and Transformer are context-agnostic baselines and others are context-aware baselines. The results with the symbol ``\dag'' are directly reported from the previous work. BLEU points with the symbol ``*'' are re-implemented by ourselves. ``\ddag'' denotes our proposed method.}
\resizebox{0.95\textwidth}{!}{
\begin{tabular}{l|cccc}
\toprule
Mono-lingual Context & IWSLT-2017 & NC-2016  &  Europarl  &  WMT-2019 \\ \midrule
RNN \cite{rnnsearch} & 19.24$^{\dag}$  &16.51$^{\dag}$  &26.26$^{\dag}$ &- \\
Transformer \cite{Transformer} &23.28$^{\dag}$ &22.78$^{\dag}$ &28.72$^{\dag}$ &- \\
Transformer (our re-implementation) &24.52* &24.45* &29.98* &38.02* \\
\midrule
ECT \cite{Extended_Context} & 24.32*  &24.40*  &30.08* &38.14*\\
Dual Encoder \cite{larger_context} & 24.14*  &24.36*  &30.12* &38.12*  \\
DCL \cite{DCL}    & 24.00$^{\dag}$  &23.08$^{\dag}$  &29.32$^{\dag}$ &-        \\
HAN \cite{HAN}  & 24.58$^{\dag}$  &25.03$^{\dag}$  &28.60$^{\dag}$ &-         \\
Transformer + QCN \cite{QCN} & 24.41$^{\dag}$  &22.22$^{\dag}$  &29.48$^{\dag}$ &-        \\
SAN \cite{selective_attention} & 24.55$^{\dag}$ &24.78$^{\dag}$ &29.75$^{\dag}$ &-      \\
Flat Transformer \cite{Flat_Transformer}& 24.87$^{\dag}$  &23.55$^{\dag}$  &30.09$^{\dag}$ &38.34* \\
\midrule
\textbf{\ourmethod{} (our method)} &\textbf{24.94$^{\ddag}$} &\textbf{25.22$^{\ddag}$}  &\textbf{30.49$^{\ddag}$} &\textbf{38.52$^{\ddag}$} \\\bottomrule
\end{tabular}}

\end{center}
\end{table*}

\begin{table*}[t]
\begin{center}
\caption{\label{two-pass}  Sentence-level evaluation results on four tasks with BLEU\% metric using the bi-lingual context. Bold numbers represent the best BLEU points. The results with the symbol ``\dag'' are directly reported from the previous work. BLEU points with the symbol ``*'' are re-implemented by ourselves. ``\ddag'' represents our proposed method. }
\resizebox{0.95\textwidth}{!}{
\begin{tabular}{l|ccccc}
\toprule
Bi-lingual Context      & IWSLT-2017 & NC-2016        &  Europarl  &  WMT-2019 & \\ \midrule
ECT \cite{Extended_Context} & 24.38*  &24.55*  &30.24* &38.16*   \\
Dual Encoder \cite{larger_context}  & 24.26*  &24.46*  &30.25* &38.24*  \\
DCL \cite{DCL} &23.82$^{\dag}$  &22.78$^{\dag}$  &29.35$^{\dag}$ &- \\
HAN \cite{HAN} &24.39$^{\dag}$  &24.38$^{\dag}$  &29.58$^{\dag}$ &-      \\
CADec  \cite{CADec}  & 24.45*  & 24.30*  &29.88*  &-    \\
SAN \cite{selective_attention}& 24.62$^{\dag}$  &24.36$^{\dag}$  &29.80$^{\dag}$ &-      \\
\midrule

\textbf{\ourmethod{} (our method)} &\textbf{25.04$^{\ddag}$} &\textbf{25.28$^{\ddag}$} &\textbf{30.89$^{\ddag}$} &\textbf{38.55$^{\ddag}$} \\
\bottomrule
\end{tabular}}

\end{center}
\end{table*}

\subsection{Datasets}
To evaluate our method, we use the same dataset as previous work, including IWSLT-2017, NC-2016, Europarl, and WMT-2019 En-De translation \cite{selective_attention}.
\paragraph{IWSLT-2017} This corpus is from IWSLT-2017 MT track and contains transcripts of TED talks aligned at the sentence level. 
\paragraph{NC-2016}  NC-2016 dataset is from Commentary v9 corpus. Newstest2015 and newstest2016 are used as the valid and the test set. 
\paragraph{Europarl} The dataset from Europarl v7 is split into training, valid and test sets according to the previous work \cite{selective_attention}. Europarl is extracted from the European Parliament website. 
\paragraph{WMT-2019}
The WMT-2019 dataset comes from the WMT-2019 news translation shared task for English-German. Newstest2016, newstest2017, and newstest2018 are concatenated as the valid set. Newstest2019 is used as the test set.\footnote{\url{https://www.statmt.org/wmt19/}}

\subsection{Implementation Details}
Considering the model performance and computation cost, we use one previous and one next sentence as the source and target context for all our experiments. 
The evaluation metric is case-sensitive tokenized BLEU \cite{BLEU}. For different benchmarks, we adapt the batch size, the beam size, the length penalty, the number of unified self-attention layers $N_1$, and the number of selection layers $N_2$ to get better performance. 
For all experiments, we use a dropout of 0.1 and cross-entropy loss with a smoothing rate of 0.1 for sentence-level and context-aware baselines except notification. All sentences are tokenized with Moses \cite{Moses} and encoded by BPE \cite{BPE} with a shared vocabulary of 40K symbols. The batch size is limited to 2048 target tokens by default.
For the \textbf{IWSLT-2017} dataset, we deploy the small setting of the Transformer model, which has 6 layers with 512 embedding units, 1024 feedforward units, 4 attention heads, a dropout of 0.3, a $l_2$ weight decay of 1e-4. 
For the \textbf{NC-2016} dataset, we use the base setting of Transformer \cite{Transformer}, in which both the encoder and the decoder have 6 layers, with the embedding size of 512, feedforward size of 2048, and 8 attention heads. We set both dropout and attention dropout as 0.2 for our method. For the \textbf{Europarl} and the \textbf{WMT-2019} dataset, the base setting of the Transformer model with 4000 warming-up steps is used. 

\begin{table*}[t]
\begin{center}
\caption{Sentence-level evaluation results on four benchmarks with BLEU\% metric under the mono-lingual context setting. The architecture $N_1+N_2$ represents our \ourmethod{} consists of $N_1$ unified self-attention layers and $N_2$ selection layers. The architecture ($N_1=6$, $N_2=0$) only uses six unified self-attention layers with the segment embedding, which select all context words to generate the final translation.}
\resizebox{0.95\textwidth}{!}{
\begin{tabular}{c|ccccc}
\toprule
Architecture & IWSLT-2017 & NC-2016  &  Europarl  &  WMT-2019 &  Average \\ 
\midrule
6 + 0 &24.48          &24.64          &30.22     &  38.22     &29.39    \\
5 + 1 &24.32          &24.95          &30.52     &  38.46     &29.74    \\
4 + 2 &24.55          &24.52          &30.72     &  38.37     &29.54    \\
3 + 3 &24.64          &24.88          &30.65     &  38.12     &29.58    \\
2 + 4 &24.74          &24.60          &30.62     &  \textbf{38.62}     &29.65    \\
1 + 5 &\textbf{24.94} &\textbf{25.22} &\textbf{30.49} &38.40      &\textbf{29.85} \\
0 + 6 &24.56          &24.75          &30.66     &37.98      &29.49              \\
\bottomrule
\end{tabular}}
\label{architecture}
\end{center}
\end{table*}

\subsection{Baselines}
For the mono-lingual and the bi-lingual context setting, we compare our method with other baselines. 
\paragraph{Mono-lingual Context:} \textbf{RNN} \cite{rnnsearch} and \textbf{Transformer} \cite{Transformer} are backbone models. \textbf{ECT} \cite{Extended_Context} simply concatenates the source sentence and context into the standard Transformer model.
Besides, \textbf{Dual Encoder} \cite{larger_context} uses two encoders to incorporate the source sentence and context sentences to predict the translation. 
Moreover, \textbf{DCL} \cite{DCL} incorporates context hidden states into both the source encoder and target decoder.
\textbf{Flat Transformer} \cite{Flat_Transformer} focus on the current self-attention at the top. Furthermore, \textbf{HAN} \cite{HAN} and \textbf{
SAN} \cite{selective_attention} introduce the hierarchical and selection attention mechanism. \textbf{QCN} \cite{QCN} is a query-guided capsule networks.
\paragraph{Bi-lingual Context:}
\textbf{CADec} \cite{CADec} is composed of identical multi-head attention layers, of which the decoder has two multi-head encoder-decoder attention with encoder outputs and first-pass decoder outputs. Also, \textbf{Dual Encoder}, \textbf{ECT}, \textbf{DCL}, \textbf{HAN} and \textbf{SAN} can also use the bi-lingual context to improve the performance.

\subsection{Main Results}
\paragraph{Mono-lingual Context} We present the results of our proposed method, sentence-level baselines, and other context-aware baselines in Table \ref{one-pass}, which all only use the mono-lingual source context. The context-aware baselines include ECT, Dual Encoder, DCL, HAN, SAN, and Flat Transformer. The sentence-level Transformer model gets 24.52, 24.55, 29.98, and 38.02 BLEU points on four benchmarks. Compared to this strong baseline, our model also significantly gains an improvement of +0.42, +0.77, +0.81, and +0.51 BLEU points respectively on four benchmarks.  
Furthermore, our method outperforms SAN by +0.39, +0.44, +0.74 BLEU points on IWSLT-2017, NC-2016, and Europarl datasets. We also observe that most context-aware models gain better performance than the sentence-level model Transformer, especially on IWSLT-2017, NC-2016, and Europarl datasets. We conjecture these three datasets are suitable for evaluating context-aware models, where the current sentence needs to learn longer dependencies.

\paragraph{Bi-lingual Context} Under the bi-lingual context setting, our method also outperforms other baselines, including Dual Encoder, ECT, DCL HAN, CADec, and SAN. \ourmethod{} can achieve improvements of +1.02, +1.33, +0.91, +0.53 BLEU points than the sentence-level Transformer baseline. It proves that \ourmethod{} also can be compatible with the target context to select useful words. Besides, \ourmethod{} can significantly outperform the related baseline SAN model by +0.49, +0.50, +1.14 BLEU points, achieving better performance on three benchmarks.  
We also observe that the bi-lingual context provides marginal improvements over the mono-lingual context. According to these results, we infer that whether the context-aware model benefits from the bi-lingual context setting is dependent on the specific dataset.



\section{Analysis}
\begin{figure}[t]
\begin{center}
    \includegraphics[width=0.7\textwidth]{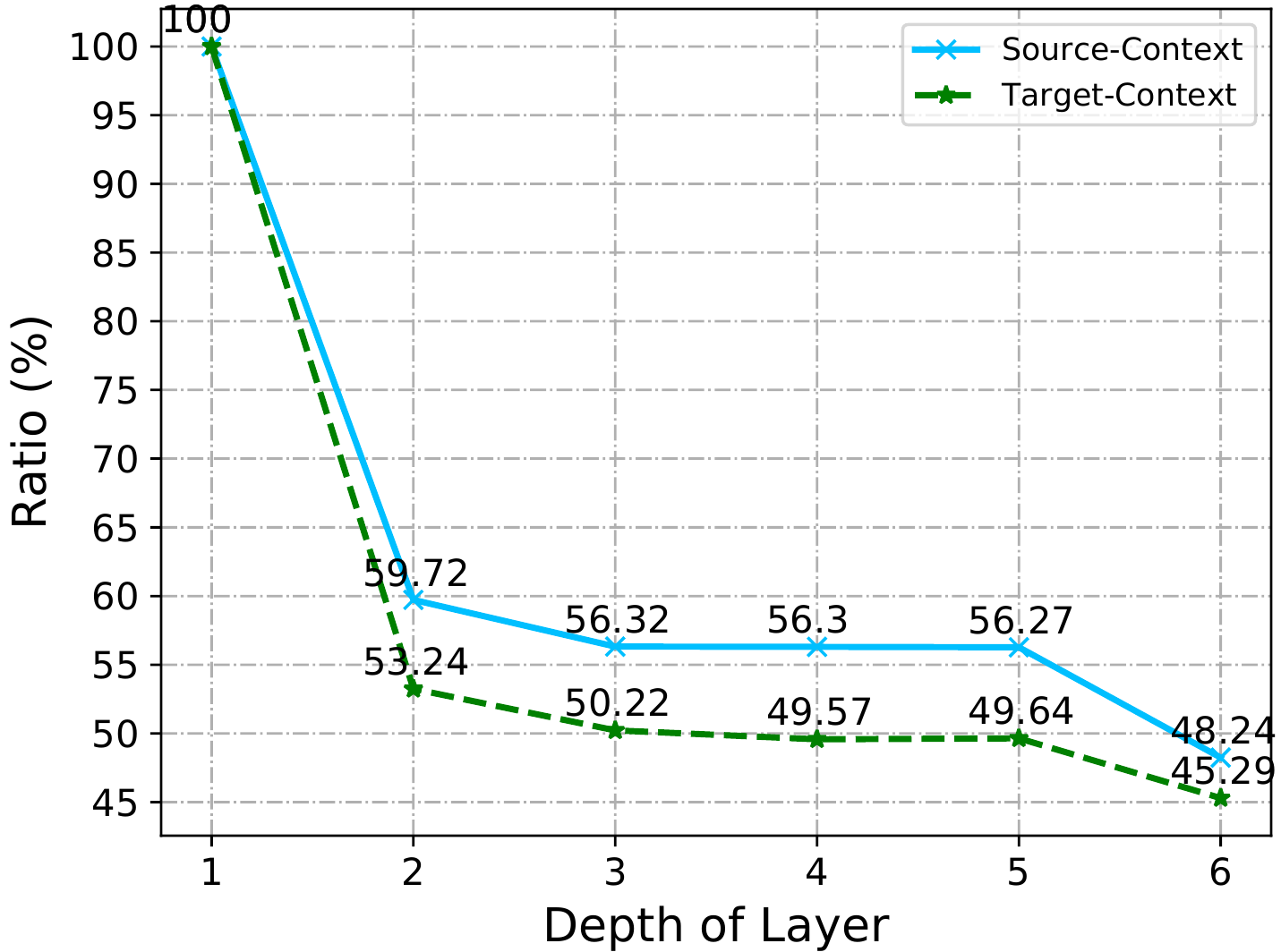}
    \caption{Ratio of the source and target context selected words on the NC-2016 dataset. Our model selects useful words from both the source and the target context gradually layer by layer. Therefore, the number of context words reduces as the depth of selection layer increasing.}
    \label{select-ratio}
\end{center}
\end{figure}
\begin{figure}[t]
\begin{center}
    \includegraphics[width=0.7\textwidth]{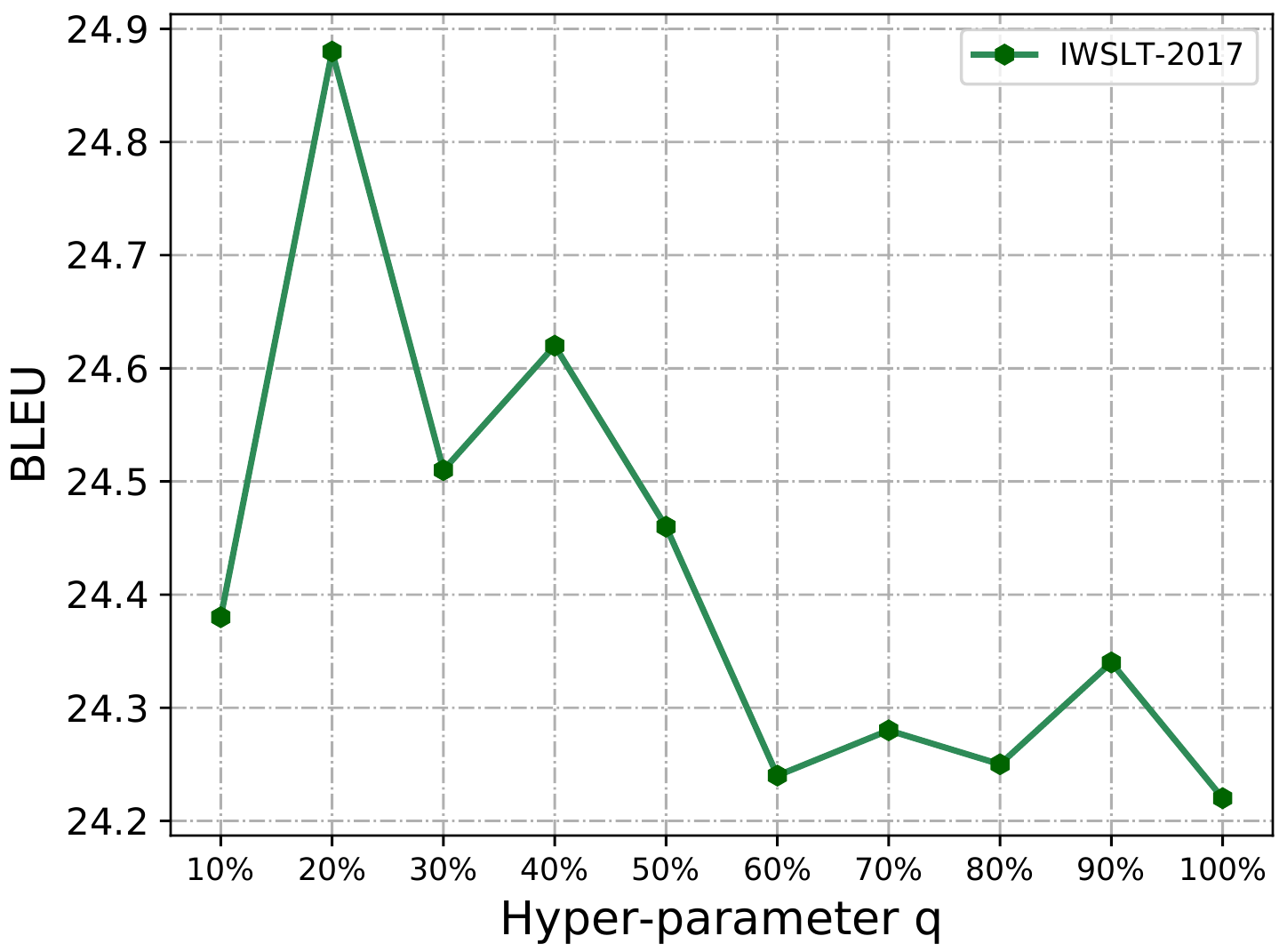}
    \caption{Results of \ourmethod{} ($N_1=5$, $N_2=1$) with different values of the hyper-parameter $q$ to control the selection of context words on IWSLT-2017 dataset.}
    \label{vote}
\end{center}
\end{figure}

\begin{figure}[t]
\centering
    \subfigure[]{
    \includegraphics[width=0.7\textwidth]{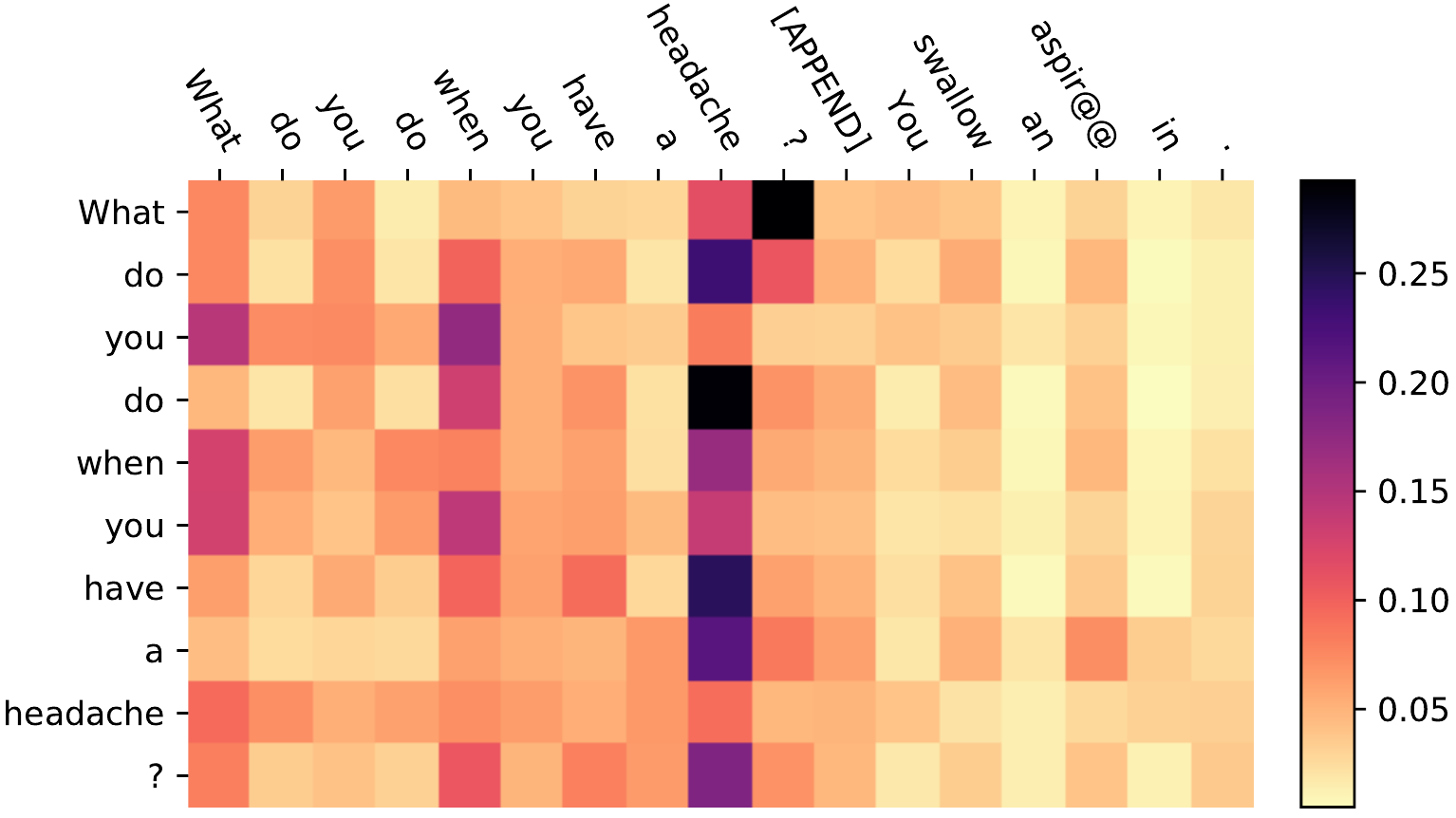}\quad
    \label{selection_attention1}
    }
    \subfigure[]{
    \includegraphics[width=0.7\textwidth]{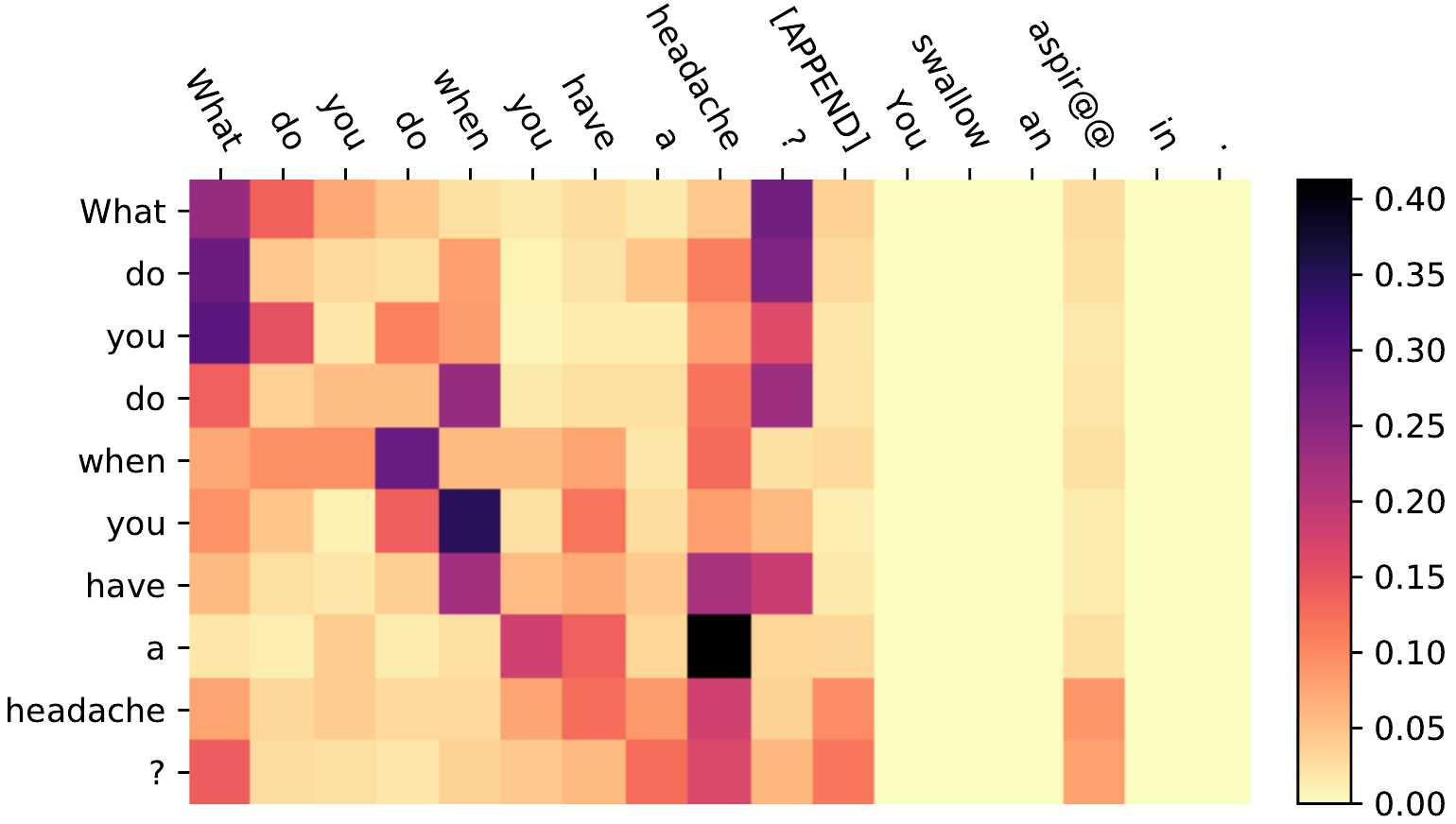}\quad
    \label{selection_attention2}
    }
    \caption{Attention visualization of the encoder self-attention weights of the bottom unified attention layer (a) and top layer after the selection operation (b). } 
\end{figure}
\label{attnetion}

\paragraph{Attention Visualization} Our model encodes the concatenation of the source words and all context words by the unified attention layers at the bottom layers. As shown in Figure \ref{selection_attention1}, the model focuses on the source sentence ``What do you do when you have a headache ?'' and all context words ``You swallow an aspir@@ in .'' using the self-attention mechanism, which ensures that all context words can provide the external guidance and implicitly contribute to the translation. The context words with higher attention weights tend to be selected.
In Figure \ref{selection_attention2}, the model only focuses on the source sentence and selected context word ``aspir@@''. The source word ``headache'' has a correlation with the context word ``aspir@@''. In this way, our method pays more attention to the current sentences and the selected words, while the other context words also provide the supplemental semantics for the current sentences at the bottom.

\paragraph{Number of Selection Layers} To better understand the impact of the selection layers for the translation performance, we tune different numbers of concatenation self-attention layers ($N_1$ layers) and selection layers ($N_2$ layers) to get the better performance under the mono-lingual setting. For the fair comparison, we keep the $N_1+N_2=6$, which equals the number of the base setting Transformer layers. As shown in Table \ref{architecture}, we find that the architecture ``$N_1=1$, $N_2=5$'' gains the best average performance on four benchmarks. Besides, stacking too many selection layers also leads to worse performance, which may be caused by wrongly discarding too many context words. In summary, our proposed model uses all context words by unified self-attention layers, and focuses those important context words at the top of encoder blocks.


\paragraph{Ratio of Selected Words} We investigate how many words in the context are selected on the NC-2016 dataset. Figure \ref{select-ratio} shows that the first selection layer reserves only 60\% words from the context. After multiple selection layers, the ratio gradually reduces to 48.24\%.
Another obvious phenomenon is that the ratio of the target context is less than that of the source context. An intuitive explanation is that we use the source sentence words to select source or target context words, where source-source attention has a higher score compared with the source-target attention. Representations of the same language have a closer relationship than those of different languages on average \cite{CoSDA-ML}. 

\paragraph{Controlling the Context Selection} According to Equation \ref{percent}, whether a context word is selected depends on $\delta_{s_k \ge q*p}$. To investigate which value of the hyper-parameter $q$ is beneficial for the translation, we tune different values of $q$ switching from 0.1 to 1.0. In Figure \ref{vote}, the results show that when the value of $1$ is in $[0.1, 0.3]$, our proposed method gets better performance. When the value of $q$ is too large, it will make our model ignore most useful context words. Therefore, we recommend the value range $q = q_0 \in [0.1, 0.3]$.

\begin{table}[t]
\centering
\caption{\label{Random} Results of our method with the offline (1 previous + 1 next) and the online (1 previous) setting.}
\resizebox{0.8\textwidth}{!}{
\begin{tabular}{l|cc|c}
\toprule
Context & IWSLT-2017 & NC-2016 & Avg. \\
\midrule
Online (1 previous) & 24.62 & 24.78 & 24.70 \\ 
Offline (1 previous + 1 next) & \textbf{24.94} & \textbf{25.12} & 25.07 \\
\bottomrule
\end{tabular}
}
\end{table}
\paragraph{Online vs. Offline Setting} Table \ref{Random} lists results of our method with different context settings. 
``1 previous'' denotes the online setting where the context only includes one previous sentence. ``1 previous + 1 next'' denotes the offline setting where the context includes one previous and one next sentence. From the table, we can find that our proposed method has the similar performance with online and offline settings on the IWSLT-2017 and NC-2016 datasets. We also try the longer context including ``2 previous + 2 next'' and ``3 previous + 3 next'', but find no significant improvement.

\paragraph{Mono-lingual vs. Bi-lingual Context} For the source and target mono-lingual context setting, our method gets 24.94 and 24.98 BLEU points on the IWSLT-2017 dataset. Furthermore, we conduct experiments with the bi-lingual context sentences and get 25.04 BLEU points, where the target context sentences are the translation of the source context sentences. The bi-lingual context setting of our method has limited improvement over the mono-lingual context setting. The reason is that the target-side context shares similar information to the source-side, which also has been found by the previous work \cite{selective_attention}.

\paragraph{Leveraging Pre-trained Model} Since the parameters of our model are the same as the standard Transformer, our model can be initialized with the pre-trained model to enhance our method. The pre-trained model BART-large \cite{bart} is used for initialization under the mono-lingual context setting. We extract 12 bottom layers of the BART encoder and 6 bottom layers of the BART decoder to initialize our model. On IWSLT-2017 dataset, our model gains +1.91 BLEU improvement (24.94 $\to$ 26.85) with pre-trained model BART.

\section{Related Work}
\paragraph{Sentence-level Machine Translation}
Sentence-level neural machine translation has developed immensely in the past few years, from RNN-based \cite{Seq2Seq,Seq2Seq2,GNMT,deliberation_network,multipass_decoder,soft_template}, CNN-based \cite{ConvS2S}, to the self-attention-based architecture \cite{Transformer,LightConv,insertion_transformer,select_relevant_nmt,UM4,HLT_MT,wmt2021_microsoft,GanLM}. However, these models always performed in a sentence-by-sentence manner, ignoring the long-distance dependencies. The past or future context can be important when it refers to using discourse features to translate the source sentence to the target sentence.

\paragraph{Context-aware Machine Translation}
Context-aware machine translation aims to incorporate the source
or target context to help translation. Previous works \cite{coherence_1,larger_context,Cross_sentence_context,Sense_Disambiguation,Evaluating_Discourse_Phenomena,Anaphora_Resolution,document_memory_networks,CADec} have proven the importance of context in capturing different types of discourse phenomena such as Deixis, Ellipsis, and Lexical Cohesion. Others \cite{larger_context,Cross_sentence_context,Discourse_phenomena,Anaphora_Resolution,selective_attention} explore the dual encoders and concatenation-based context-aware models.

Recently, a promising line of research to improve the performance of the context-aware NMT is to select useful words of the whole context, which can be used to enhance the positive use of context \cite{Document_Survey,discriminate_noise,context_aware_regularisation}.
Other researchers propose selective attention mechanism by introducing sparsemax function \cite{selective_attention,sparsemax}.

\section{Conclusion}
\label{Conclusion}
In this work, we explore the solution to select useful words from the context. We propose a novel model called \ourmethod{}, consisting of unified self-attention layers and selection layers.
The experiments on both mono-lingual and bi-lingual context settings further prove the effectiveness of our method
Experimental results demonstrate that our proposed method can select useful words to yield better performance.

%
%
\bibliography{HanoiT}
\bibliographystyle{splncs04}

\end{CJK*}
\end{document}